\documentclass[letterpaper, 10 pt, conference]{ieeeconf}  % Comment this line out if you need a4paper

\IEEEoverridecommandlockouts                              % This command is only needed if 
\usepackage{amsmath} % assumes amsmath package installed
\usepackage{amssymb}  % assumes amsmath package installed
\usepackage{graphicx} % for pdf, bitmapped graphics files
\usepackage{epstopdf}
\usepackage{bm}
\usepackage{subfigure}
\usepackage{url}

\newcommand{\argmin}{\mathop{\rm arg~min}\limits}

\title{\LARGE \bf
Development of a Stereo-vision based High-throughput Robotic System \\ for Mouse Tail Vein Injection
}

\author{Tianyi Ko$^{*1}$, Koichi Nishiwaki$^{1}$, Koji Terada$^{1}$, Yusuke Tanaka$^{1}$, Shun Mitsumata$^{2}$, Ryuichi Katagiri$^{2}$, \\ 
Junko Taketo$^{2}$, Naoshi Horiba$^{2}$, Hideyoshi Igata$^{1}$, Kazue Mizuno$^{*1}$
\thanks{$^{*}$Corresponding author.
    {\tt\small \{tko, mizuno\}@preferred.jp}
}
\thanks{$^{1}$Preferred Networks Inc., 3F Otemachi-building, 1-6-1 Otemachi, Chiyoda-Ku, Tokyo, Japan.
}
\thanks{$^{2}$Chugai Pharmaceutical Co., Ltd., Fuji Gotemba Research Labs, 1-135 Komakado, Gotemba, Shizuoka, Japan.
}
}% <-this stops

\begin{document}

\maketitle
\thispagestyle{empty}
\pagestyle{empty}

%%%%%%%%%%%%%%%%%%%%%%%%%%%%%%%%%%%%%%%%%%%%%%%%%%%%%%%%%%%%%%%%%%%%%%%%%%%%%%%%

\begin{abstract}
In this paper, we present a robotic device for mouse tail vein injection.
We propose a mouse holding mechanism to realize vein injection without anesthetizing the mouse, which consists of a tourniquet, vacuum port, and adaptive tail-end fixture.
The position of the target vein in 3D space is reconstructed from a high-resolution stereo vision.
The vein is detected by a simple but robust vein line detector.
Thanks to the proposed two-staged calibration process, the total time for the injection process is limited to 1.5 minutes, despite that the position of needle and tail vein varies for each trial.
We performed an injection experiment targeting 40 mice and succeeded to inject saline to 37 of them, resulting 92.5\% success ratio.

\end{abstract}

\section{Introduction}
Animal experiment plays important roles in various domains such as biochemistry and drug discovery.
Among the animals, laboratory mice are widely used due to the ease of handling.
Tail vein injection, or intravenous administration, is a common route to access to their bloodstream.
%Since the operation needs to be manually conducted by a skilled person, there are problems of human resource and human error.
%To maintain the skill, the technician needs a continuous training, resulting less time to spent for more creative tasks.
%To compensate for the human error between different operator, or even the same operator under different condition, additional number of mice are used.
%This is a problem in both animal welfare and cost efficiency.
%
To support the blood sample collection from mouse tail vein, Liu et. al.~\cite{liu2019modified} showed that
a combination of eyeglass and butterfly needle improves the success ratio.
Berry-Pusey et. al.~\cite{berry2013semi} developed a system that semi-automatically insert a needle into the tail vein of a mouse.
Targeting an anesthetized mouse, they used a vision-based approach to align the needle with the vein in one direction,
while for the other direction, or the depth direction, they used a pressure-sensing approach to detect whether the needle tip reaches the vein.
Chang et. al.~\cite{chang2014automated} extended the work of \cite{berry2013semi} by introducing a visual-servo controller
to replace the original man-in-the-loop one.
Fromholtz et. al.~\cite{fromholtz2017design} developed a tail vein injection system with infrared stereo vision while it lacks evaluations with real mice.

In contrast with the successful examples targeting human, such as the works by Chen et. al.~\cite{chen2020deep,chen2013portable} and Balter et. al.~\cite{balter2015system}, there are different challenges specific for mouse.
A major one is the size: the mouse tail vein is limited to around 300$\mu$m in diameter, which is around 1/10 the case in human arm.
The small scale not only requires high system accuracy, but also limits the sensing approach, e.g., ultrasound echo is effective to detect the depth of human's vein under the skin but it is difficult to be applied for mouse due to its limited resolution.
Another problem specific for animals is that since they do not understand the operator's command, we need to physically constraint their motion during the injection process.
While anesthetizing them is a solution, it makes the overall process complicated and time-consuming.
Another difficulty is that the anesthetic may biologically or chemically affect the experimental results.

\begin{figure}[t]
    \begin{center}
        \includegraphics[width=1.0\columnwidth]{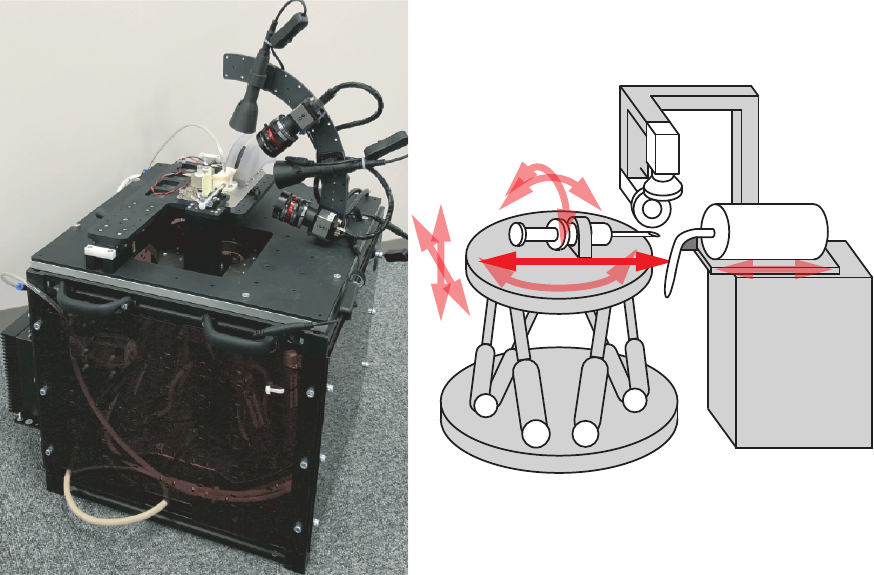}
        \caption{
            Appearance of the developed robotic mouse tail vein injection system and its schematic.
            The system has 450 mm $\times$ 450 mm footprint and 650 mm height.
            A hexapod stage inside the box-shaped basement moves the injector.
            Two cameras supported by the vertical stand detects 3D position of the target vein.
        }
        \label{fig:overall_appearance}
    \end{center}
\end{figure}

In this paper, we present a new approach to realize a robotic mouse tail vein injection system.
We propose a mouse holding mechanism to realize vein injection without anesthetizing the mouse.
To detect the 3D position of the target vein, we developed a high resolution stereo vision system.
Calibration is a critical point in a vision-based system.
A challenge specific for tail vein injection is that the position of both the needle and the target vein differs from each trial.
We propose a two-staged calibration process to reduce the calibration time for each trial.
In the last of the paper, we present an evaluation of the developed system with 40 real mice.

The paper is organized as follows.
Section~\ref{sec:overview} provides the overview of the system architecture.
Section~\ref{sec:holder} describes the mouse holding mechanism.
The vision system and calibration process is shown in section~\ref{sec:vision}.
In section~\ref{sec:evaluation}, we report a quantitative evaluation of system accuracy in addition to an injection evaluation with 40 real mice.
Section~\ref{sec:conclusion} concludes the paper.

\section{System Overview}
\label{sec:overview}

% \begin{figure}[t]
%     \begin{center}
%         \includegraphics[width=0.8\columnwidth]{./figures/overall_schematic.pdf}
%         \caption{
%             Schematic of the mouse tail vein injection system.
%             The injector is fixed on a 6-axis hexapod and aligned with the vein based on the images acquired from the two cameras.
%             The two cameras are fixed against the system basement.
%         }
%         \label{fig:overall_schematic}
%     \end{center}
% \end{figure}

Figure~\ref{fig:overall_appearance} shows the appearance of the developed robotic mouse tail vein injection system and its schematic.
The system has 450 mm $\times$ 450 mm footprint and 650 mm height.
A hexapod stage inside the box-shaped basement moves the injector.
Two cameras supported by a stand detects 3D position of the target vein.
The mouse is held inside a mouse holding mechanism, which is attached on a linear stage.
The linear stage is for the setup process such as attach/detach of mouse/injector and calibration.

\begin{figure}[t]
    \begin{center}
        \includegraphics[width=0.9\columnwidth]{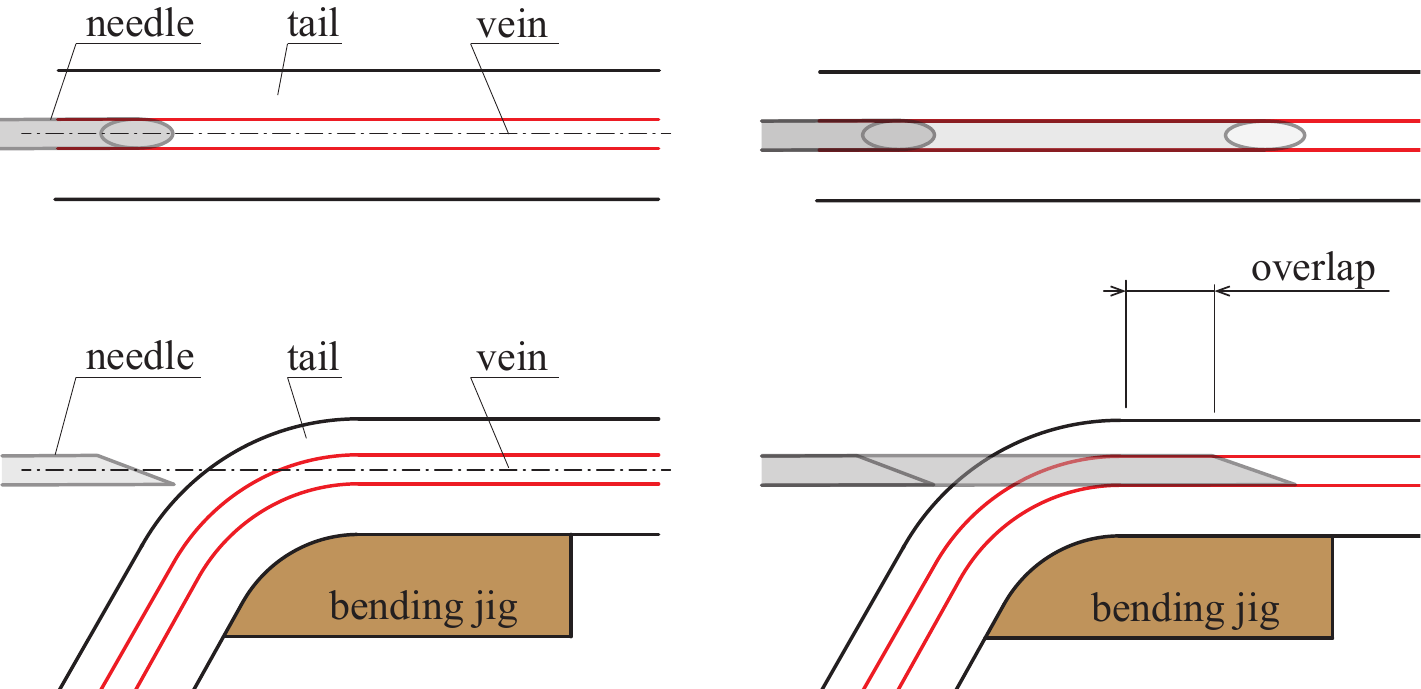}
        \caption{
            Schematic of the needle approach process.
            In the figure, the right side directs to the base of the tail and the left side directs to the tip.
            The tail is fixed on a bending jig, which consists of a bending part and a straight part.
            The needle is aligned with the vein at the straight part in the 3D space before the insertion.
            During the insertion, the needle keeps to be aligned with its initial pose.
            The needle approaches to the vein through the bent part.
        }
        \label{fig:tail_schematic}
    \end{center}
\end{figure}

One challenging requirement of tail vein injection is the need of overlap of the needle with the vein.
It is not enough to simply make the needle tip to reach inside the vein because the needle is easy to be ejected from the vein due to the mouse's motion or internal pressure.
In the case of human technician, after the needle approaches and reaches to the vein with a certain angle, the needle is tilted with its tip as the rotational center.
After the needle is tilted to be parallel with the vein, the operator insert the needle a few millimeters more to make the overlap.
In the works of \cite{berry2013semi, chang2014automated}, the requirement was relaxed by anesthetizing the mouse.
In the work of \cite{fromholtz2017design}, this overlap is not considered.
In our case, we took an approach as illustrated in Fig.~\ref{fig:tail_schematic}.
The tail is fixed on a bending jig, which consists of a bending part and a straight part.
The needle is aligned with the vein at the straight part in the 3D space before the insertion.
During the insertion, the needle is kept aligned with its initial pose.
The needle approaches to the vein through the bent part.
Since the needle and the vein is aligned, even after the needle reaches to the vein, the needle can proceed more to realize the overlap.
While in the works of \cite{berry2013semi, chang2014automated} they also bent the tail, they did not align the needle with the vein in 3D space.
Since the needle is only aligned in one direction, it is not guaranteed that the overlap is possible.

While a 4-axis stage is used for \cite{berry2013semi, chang2014automated} and a 3-axis system was developed in \cite{fromholtz2017design}, at least 5 DoF is needed to align the needle with the straight part of the vein in 3D space because the tissue's compliance leaves no guarantee that we can constraint the vein to be strictly parallel to any axis.
Another reason that we require a 5 DoF system is to compensate for the mechanical error of the injector.
Since experiments with vein injection commonly use a large number of mice, operation speed is a critical factor to decide whether the system is acceptable in the real applications.
To reduce the operation time, the developed system supports all-in-one injectors, i.e., the needle is directly attached to the syringe, so that the injector is replaced to a new one with a one-touch snap fit.
Unlike the separate types where the needle and syringe are separated and connected with a tube, it is difficult to mechanically constraint needle's pose of an all-in-one injector, thus the misalignment between the needle and syringe holder needs to be compensated by the motion.
In this work, instead of building a custom-made 5-axis stage, we adopt an off-the-shelf 6-axis hexapod (PI H-840) and leave one DoF constant.

\section{Mouse Holding Mechanism}
\label{sec:holder}

\begin{figure}[t]
    \begin{center}
        \includegraphics[width=0.9\columnwidth]{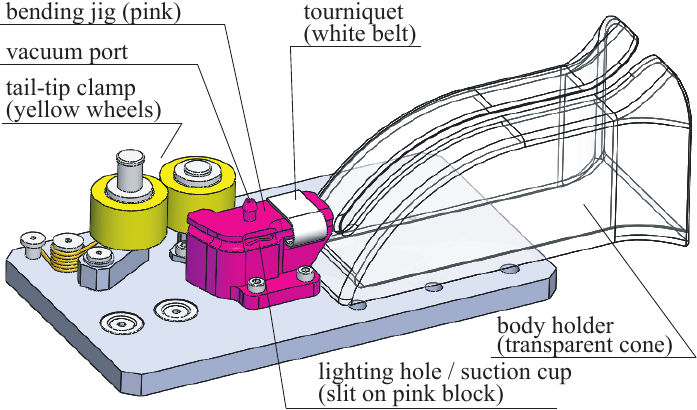}
        \caption{
            Schematic of the mouse holder assembly.
            The mouse holder has a shape to avoid the mouse from kicking the sidewall.
            The bending jig has a lighting hole to illuminate the tail.
            The lighting hole also works as a suction cup to maintain the tail to be aligned with the straight part of the jig.
            A tourniquet is attached to the bending jig to avoid tail motion and to maintain the visibility of the vein.
            Two rollers supported by a one-way clutch hold the tip of the tail.
        }
        \label{fig:mouse_holder}
    \end{center}
\end{figure}

Mouse holding mechanism plays a critical role in a system targeting non-anesthetized mice.
Figure~\ref{fig:mouse_holder} illustrates its structure, which consists of a mouse body holder, tourniquet, tail bending jig, and tail-tip holder.
The body holder is a square-shaped tube with a slit for the tail.
It has a cone-shaped entrance to avoid the mouse to hang on the edge during the setup process.
The hip side end of the holder has a diagonal shape, holding the hip by the diagonal wall while avoiding the mouse's hind leg to kick it.
% The body holder has no mechanism to hold the mouse's head.
% Mouse holders used for manual operation commonly has an additional part, or head holder, to be set into the mouse holder after setting the mouse.
% While such mechanism compress the mouse between the head holder and the hip side of the 

\begin{figure}[t]
    \begin{center}
        \includegraphics[width=1.0\columnwidth]{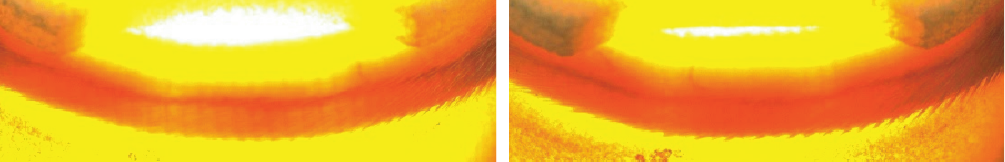}
        \caption{
            View of the tail vein from the upper main camera.
            The red dark line on the tail surface is the target vein.
            The left picture is taken right after clamping the tail, where the target vein is clear.
            The right picture is taken with the same setup but 3 minutes later.
            The vein becomes less clear since the clamp prohibits the blood stream.
            This problem is solved after adding a tourniquet.
        }
        \label{fig:without_torquinet}
    \end{center}
\end{figure}

A tourniquet is integrated into the tail bending jig and maintains the vasodilation of the vein.
Without the tourniquet, we experimentally confirmed that the vein narrows and becomes inaccessible rapidly after the mouse is set on the system.
This is due to the relatively strong force applied to the tail-tip by the tail holding mechanism, since it needs to support the strong force of non-anesthetized mice.
Figure~\ref{fig:without_torquinet} shows the view of the tail without the tourniquet, where the vein with high visibility (left image) disappears in 3 minutes (right image.)
The tourniquet also plays an important role to constraint the tail's rotational motion around its longitudinal axis.
In this work, we keep the tourniquet attached on the tail through the whole injection process since we confirmed that it does not prohibit us from injecting saline into the mouse.

\begin{figure}[t]
    \begin{center}
        \includegraphics[width=1.0\columnwidth]{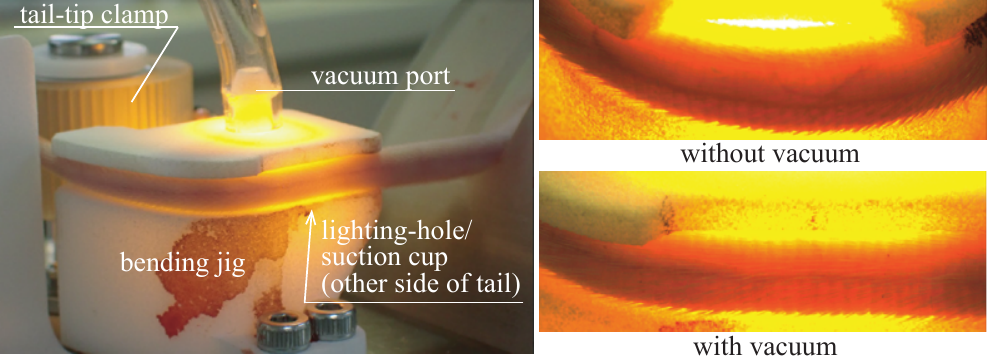}
        \caption{
            Enlarged view of the tail sticking on the bending jig with the vacuum (left) and a comparison of the tail condition seen from the embedded camera (right).
            Without the vacuum, the tail does not fit to the straight part of the jig due to its stiffness.
            In the case with the vacuum, the tail and vein is straight so that we can align the needle with it.
        }
        \label{fig:vacuum}
    \end{center}
\end{figure}

The bending jig has a slit to let the backlight and vacuum pass through.
%The light is to make the vein detection easier, on which we discuss later in section~\ref{sec:vision}.
The vacuum channel forces the tail to follow the jig by the vacuum.
As illustrated in Fig.~\ref{fig:tail_schematic}, the tail needs to be straight at the puncture point.
However, the real tail does not follow the straight part of the jig due to its stiffness.
To press the tail onto the jig, we applied vacuum to the slit.
Figure~\ref{fig:vacuum} show the bending jig with vacuum.
%The vacuum is applied through the transparent polyurethane tube and the tail is tightly fixed on the jig.
The right top is the view where no vacuum is applied.
The tail curves due to its stiffness thus the puncture is impossible.
On the right bottom is the case with vacuum, where the vein is straight enough for the puncture.

\begin{figure}[t]
    \begin{center}
        \includegraphics[width=0.9\columnwidth]{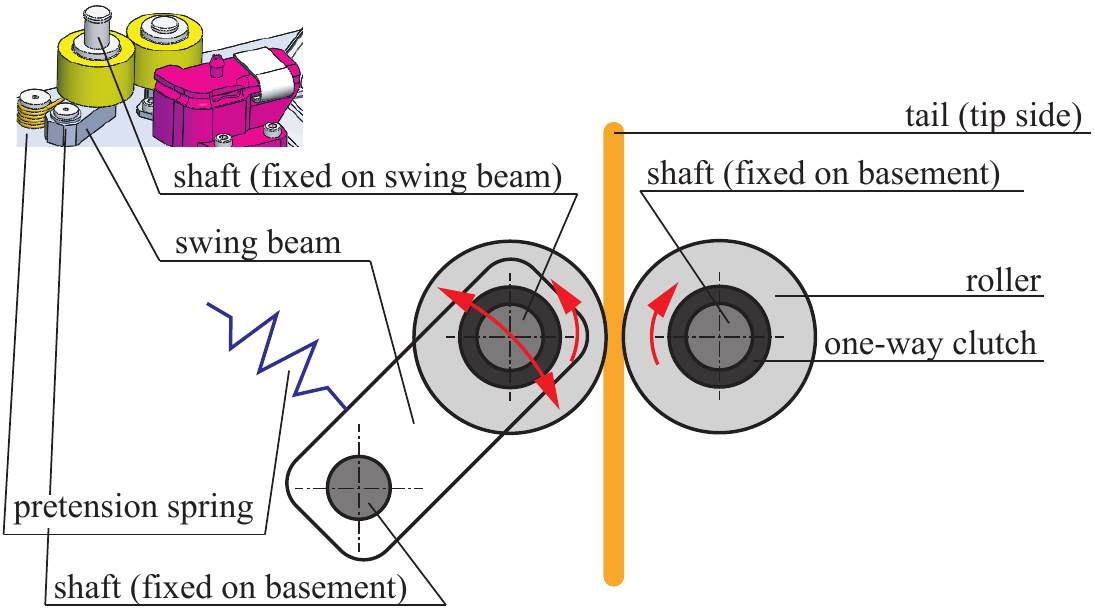}
        \caption{
            Schematic of the tail-tip holder.
            One of the roller-clutch sets is fixed on the basement, while the other is fixed on a swing beam.
            The swing beam rotates around its supporting shaft.
            It is arranged so that the more the mouse pull the tail, the stronger the wheels pinch the tail.
            The preload is generated by a spring set between the swing beam and the basement.
        }
        \label{fig:tail_clamp}
    \end{center}
\end{figure}

The tail-tip holder, illustrated in yellow in Fig.~\ref{fig:mouse_holder}, supports the mouse's force to evacuate from the system.
Two polyurethane rubber rollers are used to avoid damaging the tail.
The rollers are supported by one-way clutches, which are further supported by a spring-loaded mechanism with a proper preload.
One tradeoff on the preload is that strong preload makes it more likely to prohibit the bloodstream while weak preload makes it easier for the mouse to evacuate.
To adaptively control the preload, we introduced a design illustrated in Fig.~\ref{fig:tail_clamp}.
One of the roller-clutch set is fixed on the basement, while the other is fixed on a swing beam.
The swing beam rotates around its supporting shaft.
The preload is generated by a spring set between the swing beam and the basement.
The key is the position of the beam's shaft: it is arranged so that the more the mouse pull the tail, the stronger the wheels pinch the tail.

\section{Vision System and Calibration}
\label{sec:vision}

\subsection{Vein Detection}
\label{sec:vein_detection}
The vein is detected in the visible light region.
While the previous works~\cite{berry2013semi, chang2014automated, fromholtz2017design} detect the vein in the infrared region, the visible light region is more advantageous in the choice of available camera, lense, and light source.
As shown in Fig.~\ref{fig:without_torquinet} and Fig.~\ref{fig:vacuum}, the vein is easy to detect even in the visible light region since the tail is thin and the backlight is strong.
We evaluated multiple light source and selected an amber-colored one with 590 nm peak intensity.
%White-colored lights were not as effective as amber-colored ones since white-colored LED has a strong peak in the blue-colored region.
%We confirmed that the vein is less clear in the blue region compared with the one in the green and red regions.
%Amber is effective since it is visible from both the green and red channel of an RGB camera.
As the main cameras, we use iDS UI-5880CP industrial camera with 3088 $\times$ 2076 resolution.
In the workspace, one pixel is equivalent to 7 $\mu$m distance, which means that the 0.3 mm diameter of the vein is projected onto 40 pixels.
This resolution is fine enough to accurately detect the vein's position.

To extract the vein in the 2D image, we simply assume the darkest point as the target.
A vertical line scans the image and in each of the vertical line, the height of the darkest point is counted as the vein point.
Since the target is to extract a line, we apply a linear fitting to the vein points.
When the vein line is denoted as $ax+b$ in the image, the problem is written as follows:
\begin{equation}
  \argmin_{a, b} \sum_x ((ax + b) - \argmin_y I(x, y))^2
  \label{eq:vein_detection}
\end{equation}
where $I(x,y) \in \mathbb{R}$ represents the brightness of the image at the pixel [$x$, $y$].
In the actual implementation, we use RANSAC~\cite{ransac} instead of least square fitting to omit outliers originated by dirt and hair on the tail.
Since the vein is visible through both the green and red channel, we apply the fitting for each channel independently and adopt the one with less outliers.

\begin{figure}[t]
    \begin{center}
        \includegraphics[width=1.0\columnwidth]{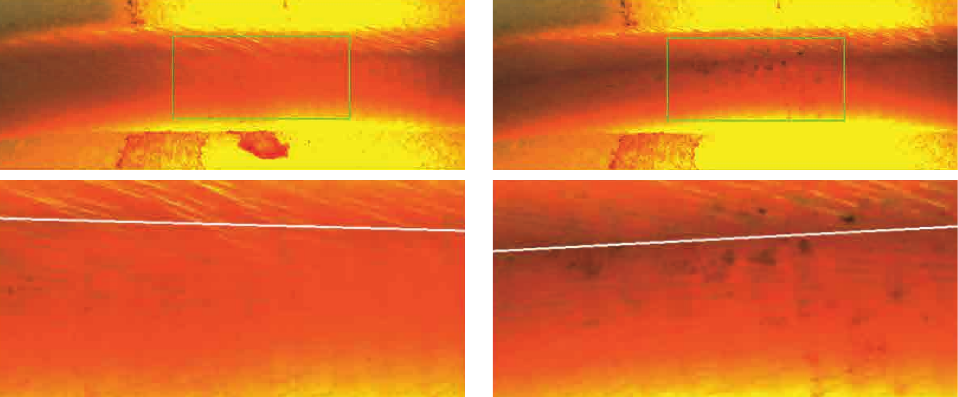}
        \caption{
            Result of the vein detector.
            The left column one is affected by the hair and the right one is affect by dirt.
            The top images are the raw images from the camera, with AoI overlaid as the rectangle.
            In the bottom image, the detected vein line is overlaid as a white line on the AoI image.
        }
        \label{fig:vein_detection}
    \end{center}
\end{figure}

Figure~\ref{fig:vein_detection} shows two examples of the vein detection.
The left column one is affected by the hair and the right one is affect by dirt.
The top images are the raw images from the camera, with the area of interest (AoI), or the target region for the vein detection, overlaid as the green rectangle.
In the bottom images, the detected vein line is overlaid as a white line on the AoI image.
The detector successfully rejects noises such as hair or dirt and properly extracts the vein.

\subsection{Needle Detection}
\begin{figure}[t]
    \begin{center}
        \includegraphics[width=1.0\columnwidth]{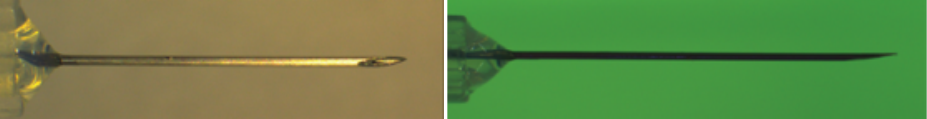}
        \caption{
            Normal view of the needle (left) and the view with the fluorescent back panel (right).
            In the normal view, the specular reflection makes the needle detection difficult.
            With the illuminous back panel, the needle tip can be easily and accurately detected from the silhouette image.
        }
        \label{fig:needle_backlight}
    \end{center}
\end{figure}

Detection of the needle position is not straight forward due to the specular reflection of the metal surface.
In such cases, placing a back light and detect the target as a silhouette is effective.
The problem in our system is that there is no sufficient space to place a back light with a uniform brightness and enough size to cover the field of view of two cameras.
To realize a thin, free form, and uniformly bright back panel, we make the panel with a fluorescent material and emit UV light from the camera side.
The specular reflection on the needle surface does not affect the camera image since it is in the UV range, while the back panel illuminates in the visible light range.
The third person view of the fluorescent back panel is shown in Fig.~\ref{fig:usbcam} and the view from the main camera is shown in Fig.~\ref{fig:needle_backlight}.
To detect the direction and tip of the needle, we used a combination of the Fast Line Detector~\cite{fld} and sub-pixel corner detector~\cite{corner_detection}.

\subsection{Two-stage Calibration}
To correctly insert the needle into the 0.3 mm-diameter vein, it is needed to accurately calibrate the system parameters such as the offset of needle tip from the hexapod's origin, internal camera parameters (focal length and distortion), and external camera parameters (pose and offset of camera from the system origin.)
The problems is that the calibration commonly requires a tedious process to gather enough number of data points.
In our case, we need to replace the injector each time after the injection to avoid contamination.
Since it is not acceptable to perform the whole calibration process for each injection due to the time constraint,
we took a two-stage manner: \textit{initial calibration} is performed once to calibrate the whole system and \textit{needle calibration} is performed before each injection to update the needle position only.

For the \textit{initial calibration},the hexapod moves the needle tip to draw a 3D grid in the work space.
For each point on the grid, the pose of the hexapod and position of the detected needle tip on the 2D image of the two cameras are recorded.
To calibrate the translation from the hexapod to the needle tip, the rotation of the needle tip varies randomly for each grid point.
In our case, the 10 mm $\times$ 2.5 mm $\times$ 2.5 mm work space is divided into a 5 $\times$ 5 $\times$ 5 grid.
The system requires around 10 minutes to gather the 125 data points.

The parameters are identified by solving the following optimization problem
\begin{equation}
    \argmin_{\bm{q}, \bm{c}_1, \bm{c}_2} \sum_{i=1}^{n}\sum_{j=1,2}\|\bm{p}_i^j - \bm{f}({\bm{H}_i}\bm{q}|\bm{c}_j)\|^2
    \label{eq:calibration}
\end{equation}
where $i=\{1, ..., n\}$ denotes the $i$-th data point (in this case $n=125$), $j=\{1, 2\}$ denotes the two cameras, $\bm{p}_i^j \in \mathbb{R}^2$ represents the position of the needle tip detected on the $j$-th camera's image in the $i$-th data point, $\bm{H}_i \in \mathbb{R}^{4\times 4}$ is the homogeneous matrix of the hexapod in $i$-th data point, $\bm{q}=[x, y, z, 1]^T$ is the translation from the hexapod to the needle tip in the hexapod frame, and $\bm{f}(\cdot|\bm{c}) \in \mathbb{R}^2$ is the projector that uses the camera parameter $\bm{c}$ to project a point in the world frame to the camera image.
The camera parameter $\bm{c}$ contains both the internal and external parameters, which is the focal length, translation, and rotation of the camera.
In this work, we do not consider the lense distortion because the field of view is narrow.
%We experimentally confirmed that in most cases the average reprojection error falls into a few pixels, which is accurate enough for our purpose taking account that one pixel is equivalent to 7 $\mu$m.

After each time the injector is replaced to a new one, \textit{needle calibration} updates the needle tip offset $\bm{q}$.
This time the camera takes only one image of the newly attached needle at the home position so that the calibration is performed instantaneously.
It is equivalent to solve the same problems as Eq.~\ref{eq:calibration} with $\bm{c}_{1, 2}$ fixed and $n=1$.
In the actual implementation the detected needle tip in the camera images are simply projected onto the 3D space by the camera parameters, instead of solving the optimization problem.
% Since not only the tip position but also the direction of the needle is required, 

\section{Experiment and Evaluation}
\label{sec:evaluation}
\subsection{Evaluation of Two-staged Calibration}
This subsection evaluates the system's accuracy and repeatability.
We first performed an \textit{initial calibration} to calibrate the whole system.
The process took 460 seconds and resulted in 0.90 pixel average re-projection error for both cameras.
Considering that one pixel is equivalent to 7 $\mu$m, the sub-pixel-level error proved that the system has enough accuracy.

To evaluate the second stage of the two-staged calibration process, we performed \textit{needle calibration} for five different needles and evaluated how well it can compensate for the needles' mechanical error.
We discuss the error in the camera image rather than in the 3D space.
This is because the error is too small to measure for an external 3D measurement system.
We denote the needle tip's position as ${}_c\bm{p}_i^j \in \mathbb{R}^2$ and needle's orientation as ${}_c\theta _i^j$ in the camera images.
The subscription $c \in \{1, 2\}$ denotes the upper/lower camera and $i \in \{0, 1, 2, 3, 4, 5\}$ is the needle's index.
Needle with id 0 is the master used for the \textit{initial calibration}.
The subscription $j \in \{\textrm{init}, \textrm{comp}\}$ denotes the two cases: the case immediately after the injector is attached, and the case after the \textit{needle calibration} is performed and the system moves the needle to the initial pose based on the detected needle pose.

\begin{figure}[t]
    \begin{center}
        \includegraphics[width=1.0\columnwidth]{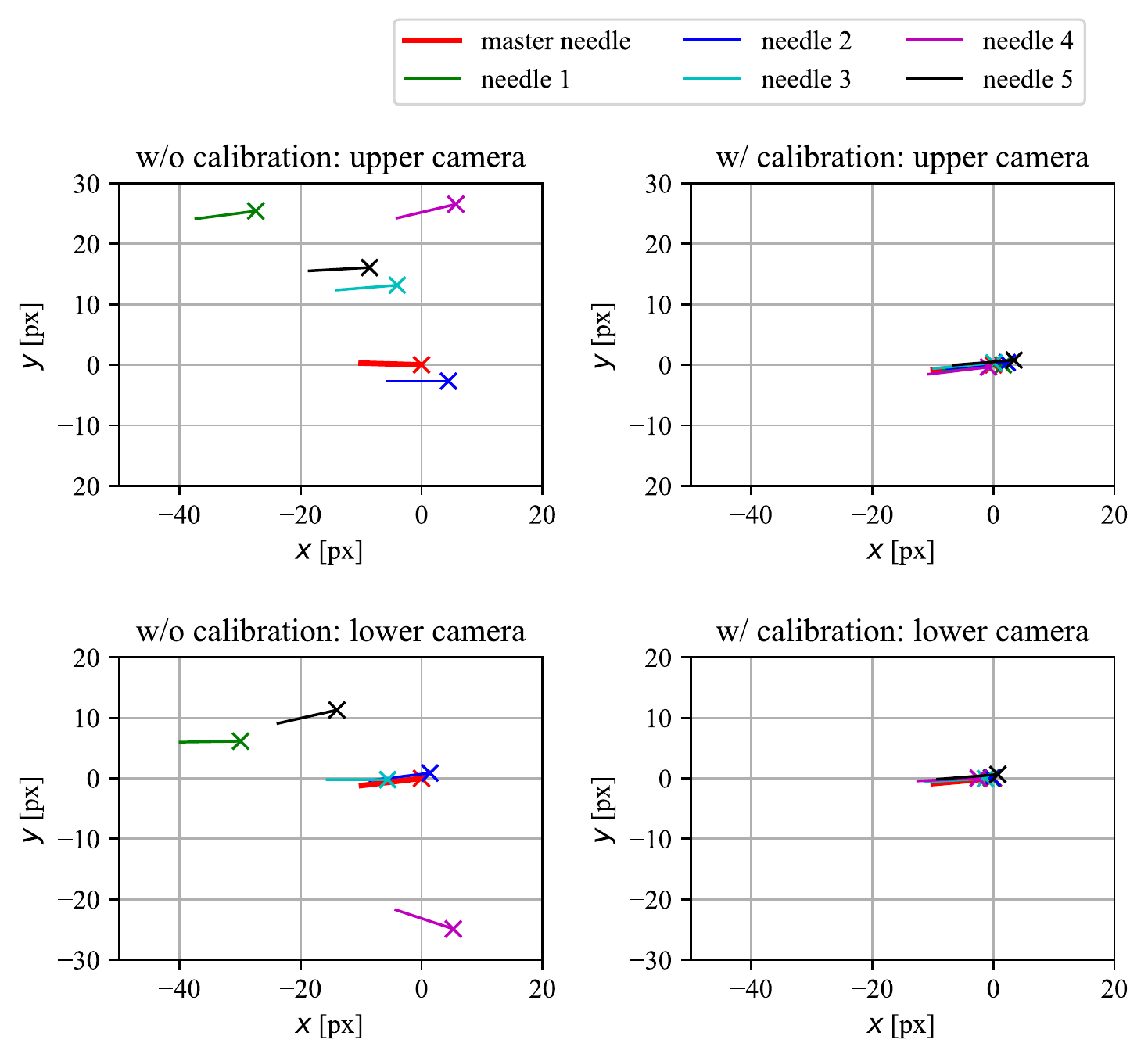}
        \caption{
            Position of the needle tip ($\times$ marker) and needle direction (line from the marker) detected on the camera images.
            The left column shows the initial condition, where the needles' mechanical variance is too large for the puncture.
            The right column shows the case after \textit{needle calibration}, where the needles' variance is compensated so that the error is small enough.
        }
        \label{fig:needle_calib}
    \end{center}
\end{figure}

Figure~\ref{fig:needle_calib} plots ${}_c\bm{p}_i^j - {}_c\bm{p}_0^j$ as $\times$ markers and ${}_c\theta_i^j$ as lines.
The camera coordinate is consistent with Fig.~\ref{fig:needle_backlight} and \ref{fig:ueye}.
Note that $\theta$ is 30 times emphasized for the reader's visibility.
The left column shows the initial case, i.e. $j=\textrm{init}$, while the right column shows the case after \textit{needle calibration} and error compensation, i.e., $j=\textrm{comp}$.
TABLE~\ref{table:calibration} summarizes the needle-tip position error $||{}_c\bm{p}_i^j - {}_c\bm{p}_0^j||$ and the needle rotation error ${}_c\theta_i^j - {}_c\theta_0^j$.

The error in the initial condition ($j=\textrm{init}$) is due to the injectors' mechanical variance.
Considering that the 0.3 mm-thick tail vein is equivalent to 40 pixels, the injector is not accurate enough to omit calibration for each trial.
The left column of Fig.~\ref{fig:needle_calib} shows a common trend: the needles seem to be originated from a similar point despite their position error.
This is reasonable since the needle is welded or glued onto the injector (Fig.~\ref{fig:ueye} well illustrates the connection part).
The result implies that the primary source of needle position error is the connection between the needle and injector.
The result also shows that the \textit{needle calibration} adequately compensates for the needle error to the acceptable level for the target task while omitting the time-consuming whole-system calibration for each injection trial.

\begin{table}[h]
    \caption{Variance of needle tip position and orientation in case w/o and w/ \textit{needle calibration}.}
    \label{table:calibration}
    \centering
    \setlength\tabcolsep{4pt}
    \begin{tabular}{cccccccccc}
        \hline
        & \multicolumn{4}{c}{w/o calibration} & & \multicolumn{4}{c}{w/ calibration} \\
        & \multicolumn{2}{c}{upper cam} & \multicolumn{2}{c}{lower cam} & & \multicolumn{2}{c}{upper cam} & \multicolumn{2}{c}{lower cam}\\
        & $p$ [px] & $\theta$ [$^{\circ}$] & $p$ [px] & $\theta$ [$^{\circ}$] & &  $p$ [px] & $\theta$ [$^{\circ}$] & $p$ [px] & $\theta$ [$^{\circ}$] \\
        \hline
        1 & 37.4 & 0.30 & 30.5 & -0.20 & & 1.6 & 0.01 & 0.3 & -0.06 \\
        2 & 5.2 & 0.05 & 1.6 & 0.01 & & 2.3 & 0.02 & 0.2 & -0.07 \\
        3 & 13.8 & 0.21 & 5.6 & -0.23 & & 0.3 & 0.01 & 1.3 & -0.06 \\
        4 & 27.2 & 0.50 & 25.4 & -0.84 & & 0.9 & 0.06 & 2.5 & -0.08 \\
        5 & 18.2 & 0.16 & 18.0 & 0.20 & & 3.5 & -0.01 & 1.0 & -0.02 \\
        RMS & 23.2 & 0.29 & 19.7 & 0.41 & & 2.1 & 0.03 & 1.3 & 0.06 \\
        \hline
        \end{tabular}
\end{table}

\subsection{In Vivo Injection Evaluation}
\begin{figure}[t]
    \begin{center}
        \includegraphics[width=1.0\columnwidth]{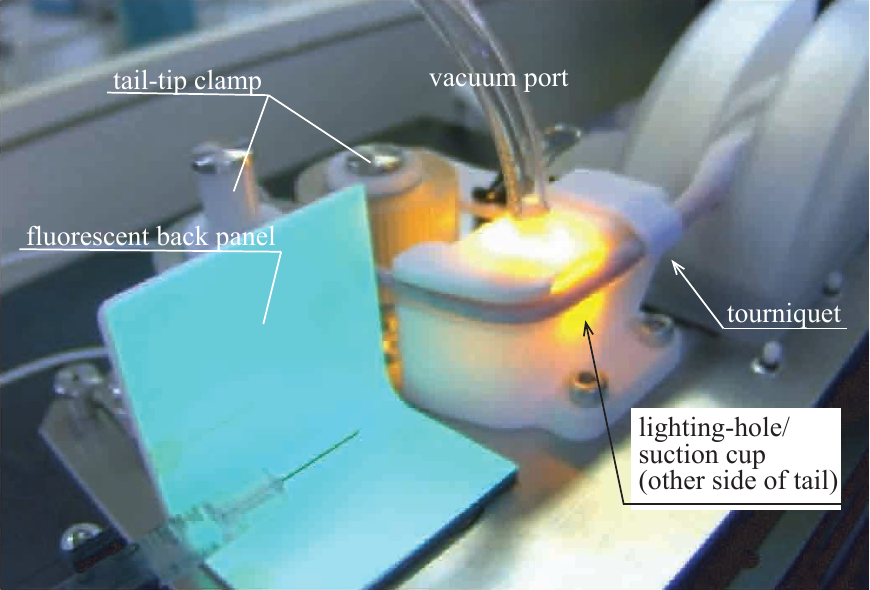}
        \caption{
            View of the injection experiment.
            On the right top, the mouse is fixed in the holder.
            The tail bending jig is in the middle of the image.
            The transparent tube attached on the bending jig is for the vacuum.
            The bright part of the jig is due to the internal LED.
            The white rubber band between the jig and the mouse is the tourniquet.
            The injector is On the left bottom.
            The L-shaped panel between the injector and the tail-tip clamp is made of fluorescent material.
        }
        \label{fig:usbcam}
    \end{center}
\end{figure}

\begin{figure}[t]
    \begin{center}
        \includegraphics[width=1.0\columnwidth]{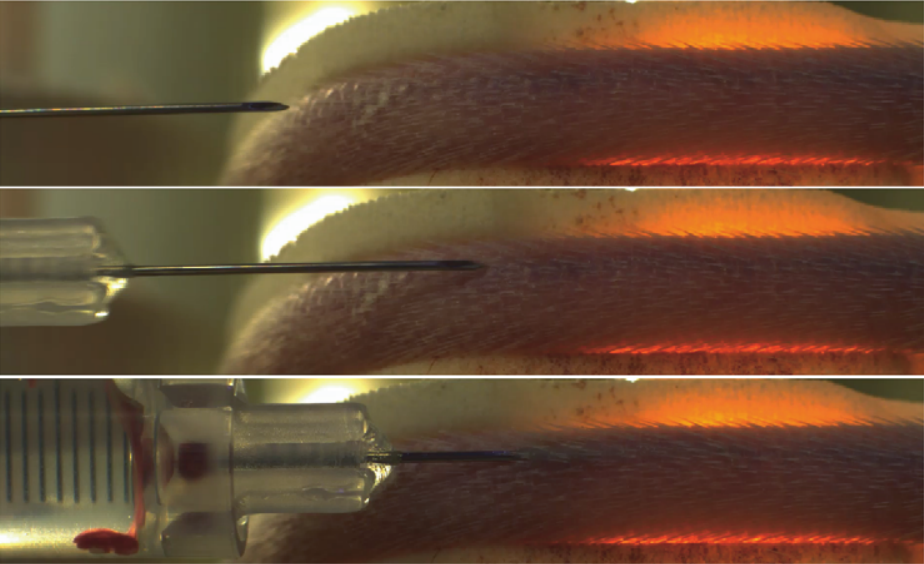}
        \caption{
            Image of the lower embedded camera during the injection experiment.
            Top: the needle and tail are at the initial position.
            Middle: the needle is aligned with the vein and ready for puncture.
            Notice that not only the position but also the rotation of the needle is changed from the initial position to compensate for the misalignment of the vein.
            Bottom: the puncture is successful and the blood flows back to the syringe.
        }
        \label{fig:ueye}
    \end{center}
\end{figure}

To evaluate the system, we performed an in vivo injection experiment.
\footnote{Animal procedures and protocols were in accordance with the Guidelines for the Care and Use of Laboratory Animals at Chugai Pharmaceutical Co. Ltd., accredited by AAALAC and approved by the Institutional Animal Care and Use Committee (IACUC approval No. 20-267).}
In order to control the variance of tail condition, we only used 9-week-old ICR~\cite{chia2005origins} mice.
Figure~\ref{fig:usbcam} shows the view of the onboard monitoring camera;
Figure~\ref{fig:ueye} shows the images from the (lower) main camera during the experiment.
The top shows the initial state; the middle shows when the needle is aligned with the vein and ready for puncture.
At this moment the needle has no contact with the tail yet.
Not only the position but also the direction of the needle varies from the initial state.
The bottom shows the view after the puncture is done.
When the puncture is successful, we can see the blood flow back into the injector.
%In the figure, the right top is the mouse in the holder.
%The tail bending jig is in the middle.
%The transparent tube attached on the bending jig is for the vacuum.
%The bright part of the jig is due to the internal LED.
%The white rubber band between the jig and the mouse is the tourniquet.
%On the left bottom is the injector filled with saline.
%The injector has a 29G needle.
%The L-shaped panel between the injector and the rubber wheel of the tail-tip holder is the fluorescent back panel to make the needle's image clear.

The success/failure is judged by a skilled technician, who performs manual IV as a daily task.
A puncture is counted as a success only when two criteria is fulfilled:
(i) the blood flows back into the injector if the piston is pulled (see the bottom of Fig.~\ref{fig:ueye}), and
(ii) the piston can be pushed with low reaction force to inject saline into the vein.
We experimentally confirmed that when the puncture fails, there is no backflow and the operator feels a strong piston reaction force to prevent injection.
TABLE~\ref{table:experiment} summarizes the result.
We performed the puncture on 40 mice and succeeded for 26 in the first trial.
Two of the 14 failed mice bled due to vein penetration.
Since our prototype currently only supports a single puncture point, the vein is no longer visible due to the bleeding.
For the rest 12 mice, on the other hand, there was no bleeding thus we performed a second try.
In the second trial, the puncture was successful for ten mice.
Regarding the rest two, one bled and one succeeded in the third try.
In total, we succeeded in injecting saline into 37 mice out of 40 (92.5\%) within three trials.
The total number of trials was 53, resulting 69.8\% success ratio against puncture.

\begin{table}[h]
    \caption{Result of the in vivo evaluation}
    \label{table:experiment}
    \centering
    \begin{tabular}{ccccc}
        \hline
        Trial & No. Puncture & No. Success & success ratio [\%]\\
        \hline
        1 & 40 & 26 & 65 \\
        2 & 12 & 10 & 83 \\
        3 & 1 & 1 & 100 \\
        Total & 53 & 37 & 69.8 \\
        \hline
    \end{tabular}
\end{table}

This result is difficult to compare with the manual operation since human operators naturally proceed and retract the needle multiple times until it reaches the vein during a single operation, i.e., it is difficult to define \textit{success rate} for human operator.
Another difference is that while human operators can select arbitrary puncture points with good vein conditions, our system currently only supports a single point on the right vein with a predefined distance from the hip.
Nevertheless, since the system's goal is to establish bloodstream access, we achieved it with a 92.5\% success ratio within three trials.
This is sufficiently high for practical use, considering that the rest 7.5\% is still possible for a human operator to perform a puncture on a different point.
Our future work includes mechanical improvements to enable puncture point adjustment and support both veins.

During the experiment, there were two major failure modes.
The most frequent failure mode was that the needle fell into the thin layer between the vein and the skin, in other words, the insertion came too shallow.
In this failure mode, there was no bleeding hence we performed a second trial after a short period (around 1 hour) to let the mouse rest.
We assume that this failure was originated from the deformation of either the tissue or the needle, which is caused by the reaction force when the needle penetrates the skin.
To further investigate this failure, a thorough investigation with different needle sizes, needle sharpness, puncture speed, and approach angle (i.e., the radius of the bending jig) is needed, which remains as our future work.
Another failure mode was that the needle penetrated the vein, or in other words, it came too deep.
In this case, the system can no longer perform another injection because the mouse bleeds in the target area.
This failure is caused by the failure of vein depth estimation.
Since the previous subsection shows that the stereo vision system has enough accuracy, the last possible error source is the vein detection.
We are currently working on an even more robust learning-based one.

\section{Conclusion}
\label{sec:conclusion}
This paper proposed a robotic mouse tail vein injection system consisting of a mouse holding mechanism, stereo vision, and a 6-axis stage.
In order to constrain the motion of a non-anesthetized mouse, we proposed a mouse holding mechanism with a tourniquet, vacuum port, and adaptively-preloaded tail-end fixture.
The tourniquet effectively constrained the rotational motion of the tail and expands the vein.
The vacuum port was effective in constraining the tail to follow the jig.
The tail-tip fixture adaptively generates a pinching force to avoid the mouse's evaluation while maintaining the bloodstream.

The needle and target vein were detected in the visible light region, and their 3D position was reconstructed in a stereo vision manner.
We proposed a simple but effective approach to detect the vein's segment.
A two-staged calibration process reduced the operation time:
\textit{initial calibration} was performed only once to fully calibrate the whole parameters while \textit{needle calibration} instantaneously updated only the needle position before each puncture.
A quantitative evaluation showed that (i) the visuomotor system had enough accuracy for the \textit{initial calibration} to result in sub-pixel level reprojection error and (ii) the \textit{needle calibration} effectively compensated the needles' mechanical variance, which is too large for the puncture without compensation.

In the in-vivo injection experiment, the system successfully established bloodstream access to 92.5\% of the targeted mice within three trials.
The rest 7.5\% mice were still possible for a human operator to perform a puncture on a different point.
The future works include mechanical improvements to enable puncture point adjustment and tuning of puncture parameters such as needle size, needle sharpness, and puncture speed.

%%%%%%%%%%%%%%%%%%%%%%%%%%%%%%%%%%%%%%%%%%%%%%%%%%%%%%%%%%%%%%%%%%%%%%%%%%%%%%%%

%\addtolength{\textheight}{-6cm}  


\bibliographystyle{IEEEtran}
\bibliography{bibs.bib}

\end{document}